\RequirePackage{fix-cm}
\documentclass[twocolumn]{svjour3}          %
\smartqed  %
\usepackage{graphicx}
\usepackage{color}
\usepackage{color,soul}
\usepackage{pbox}
\usepackage{amsmath}
\usepackage{float}
\floatstyle{plaintop}
\restylefloat{table}

\usepackage{subcaption}
\usepackage{tabularx}
\usepackage[export]{adjustbox}
\usepackage{booktabs}
\usepackage{diagbox}
\usepackage{verbatimbox}
\usepackage{multirow}
\usepackage{amssymb}
\usepackage{breakcites}
\usepackage{url}
\usepackage{hyperref}

\usepackage{array}
\newcommand{\R}{\mathbb{R}}
\newcolumntype{C}[1]{>{\centering\let\newline\\\arraybackslash\hspace{0pt}}m{#1}}

\begin{document}

\title{Asking Questions on Handwritten Document Collections}

\author{Minesh Mathew       \and
        Lluis Gomez \and
        Dimosthenis Karatzas \and 
        C V Jawahar %
}

\institute{Minesh Mathew and C V Jawahar \at
            Centre for Visual Information Technology (CVIT) - International Institute of Information Technology - Hyderabad, India \\
             Lluis Gomez and Dimosthenis Karatzas \at 
              Computer Vision Centre (CVC) - Universitat Autónoma de Barcelona - Barcelona, Spain \\
              \email{minesh.mathew@research.iiit.ac.in}      
}

\date{Received: date / Accepted: date}

\maketitle

\maketitle

\begin{abstract}

This work addresses the problem of Question Answering (QA) on handwritten document collections. Unlike typical QA and Visual Question Answering (VQA) formulations where the answer is a short text, we aim to locate a document snippet where the answer lies. The proposed approach works without recognizing the  text in the documents. 
We argue that the recognition-free approach is suitable for handwritten documents and historical collections where robust text recognition is often difficult. At the same time, for  human users, document image  snippets containing answers act as a valid alternative to textual answers. The proposed approach uses an off-the-shelf deep embedding network which can project both textual words and word images into a common sub-space. This embedding bridges the textual and visual domains and help us retrieve document snippets that potentially answer a question.
\sloppy We evaluate results of the proposed approach on two new datasets: (i) HW-SQuAD: a synthetic, handwritten document image counterpart of SQuAD1.0  dataset, and (ii) BenthamQA: a smaller set of QA pairs defined on  documents from the popular Bentham manuscripts collection. 
We also present a thorough analysis of the proposed recognition-free approach compared to a  recognition-based  approach which uses text recognized from the images using an OCR. Datasets presented in this work are available to download at  \href{http://docvqa.org}{\textcolor{blue}{docvqa.org}}

\end{abstract}

\section{Introduction}
\label{sec:intro}
\begin{figure}[tp]
    \centering
    \includegraphics[width=0.95\linewidth]{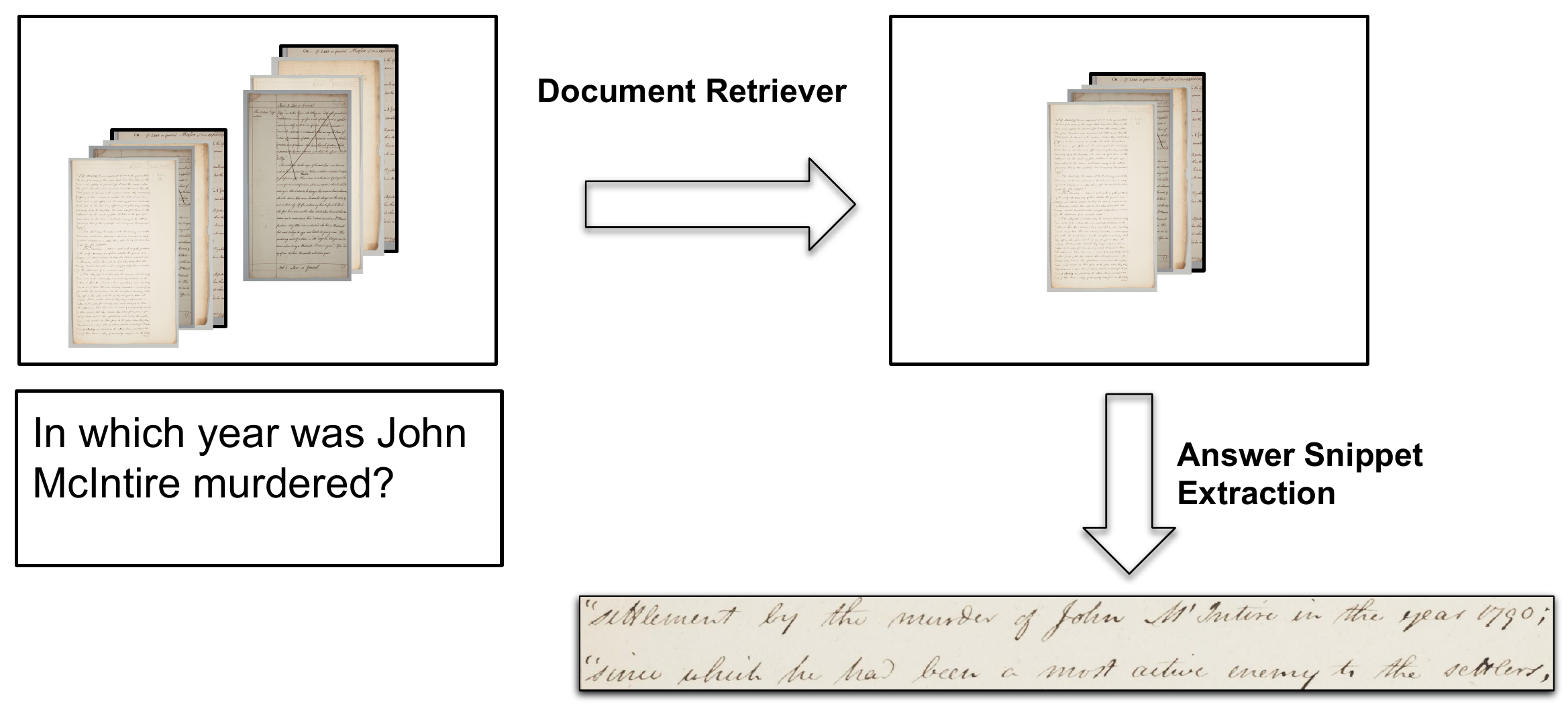}
    \caption{This work addresses the problem of Question Answering on handwritten document collection. Given a natural language question, a document snippet that answers the question is returned.}

    \label{fig:fig1:the_process}
\end{figure}

Despite the rise in the creation, use, and transmission of electronic documents, there is an increase in the information generated as paper documents.
The digitization options offered by mobile devices like phone cameras help us capture paper-based documents such as handwritten notes, medical records, or invoices, among many others
Archival and historical material are also being digitized in large quantities as part of active conservation efforts~\cite{google_books}. 
Information in such large-scale document image collections is not easy to access.
On the other hand, the ecosystem of document analysis has grown to include hand-held scanners, off-the-shelf Optical Character Recognition (OCR) tools and APIs, and cloud storage solutions. Beyond the capability of recognizing text or extract regions of interest, complex document understanding tasks have been largely left unattended by the community barring a few exceptions from recent years ~\cite{funsd,sroie,layoutlm,mathew2020docvqa}. The recent advances in deep learning  have led to significant improvements in accuracy for a wide variety of segmentation and recognition tasks~\cite{fast_cnn_layout,crnn,praveen_e2e}. It now permit us to explore more complex problems in the document comprehension space.

There are two parallel streams of work in Computer Vision (CV) and Natural Language Processing (NLP), toward measuring how well machines understand visual and textual data respectively. The computer vision community has recently defined tasks like image captioning~\cite{coco_caption_server} and Visual Question Answering (VQA)~\cite{antol2015vqa,kafle2018dvqa,textvqa,stvqa,mathew2020docvqa}. In VQA, objective performance is measured by looking at how accurately a model can answer a set of questions asked on images that humans can answer  comfortably. Similarly, the NLP and Information Retrieval (IR) communities look at understanding paragraphs and electronic text collections with Question Answering (QA) skills~\cite{squad,ms_marco,reading_wiki_acl,bert}. The questions in this space are usually semantically richer, compared to questions in VQA datasets.

In a recent work, we introduced the problem of VQA on document images and organized a challenge called DocVQA~\cite{docvqa_challenge}.
Understanding a document is a complex cognitive task that goes beyond the capability to recognize text or extract regions of interest.
DocVQA 2020 edition comprises two different tasks --- task 1 dealing with VQA on a single document image, and task 2 for VQA on a collection of forms with the same template.
Baselines~\cite{mathew2020docvqa} and the challenge results~\cite{docvqa_challenge} demonstrate that one of the  challenges in task 1 is difficulty in recognizing handwritten text.

Although deep learning based techniques have led to considerable performance improvement in handwriting recognition~\cite{praveen_e2e,hwr_eccv_2018,hwr_2020_Kang}, it is still far from OCR performance on printed documents.
Owing to large variations between collections of handwritten/historical documents, recognition models typically need to be re-trained/fine-tuned on annotated samples from the new set of documents that need to be recognized.
A recent study by Bazzo et al.~\cite{bazzo_ocr_ir} observe that even a 5\%  word error has a significant impact on information retrieval from document images that are automatically transcribed using an OCR.
In another study based on data in a large digital library, Chiron et al.~\cite{chiron_ocr_library} observe that a significant number of user queries are affected by OCR errors. They also observe that at least 15\% of the OCR errors are for named entities that cannot be corrected by dictionary-based error correction approaches. This motivates us to devise a retrieval-based solution for QA that does not require text recognition and extracts a relevant
region from an image in the collection that answers the question.

Consequently, we propose to return answers to the questions as image/document snippets.
Fig.~\ref{fig:fig1:the_process} depicts an example case where the answer to a natural language question on a document collection is an image snippet/crop from one of the documents in the collection. Although this approach does not yield a textual answer to the question, it would still meet the requirement of human users looking for information from document collections. This is in line with works in VQA which argue that visual evidence is as important as coming up with a textual answer~\cite{where_to_look,general_value_of_evidence}. For example, in the STE VQA dataset~\cite{general_value_of_evidence}, in addition to the textual answer, each question is provided with a bounding box representing the image region where the question is grounded on.
Moreover, returning image snippets as answers is similar to extractive QA tasks  in NLP\slash IR where an answer is not generated, but returned as a span (a sequence of contiguous text tokens) of the given context~\cite{squad,newsqa}.

We propose a method that performs QA in the image space, without any explicit text recognition from the documents in the collection.
We use an end-to-end deep embedding network~\cite{praveen_e2e} to extract feature embeddings of both textual words in questions and word images in documents.
These representations are aggregated for each question, document, or snippet to form question vectors, document vectors, and snippet vectors, respectively. This way, documents best matching a question or snippets best matching a question can be retrieved using the nearest neighbour search.

Major contributions of this work are: 
\begin{itemize}
    \item Formulate the problem of information extraction from a document image collection as a question answering task. This is motivated by VQA in Computer Vision and QA in NLP. The question is defined as a natural language query resulting in an answer in the form of image snippets.
    \item \sloppy Introduce two new datasets: HW-SQuAD and BenthamQA, for QA on handwritten document collection.  HW-SQuAD is curated using an existing dataset for electronic text --- SQuAD1.0~\cite{squad}. We render passages in SQuAD1.0 as images using handwritten fonts. For BenthamQA, we annotate questions and answers on handwritten manuscripts from Transcribe Bentham Project~\cite{bentham_transcribe}.
    
    \item Propose an evaluation protocol to evaluate QA, where the answer is a document image snippet.
 
    \item A two stage solution for QA on document collection and evaluate it on the newly introduced datasets. Our method is segmentation-based (requires word and line bounding boxes), recognition-free (does not require to recognize text in the documents) and lexicon-free (does not depend on a fixed vocabulary). We compare our approach against a recognition-based approach  and analyze how the two approaches fare when text recognition is noisy.
\end{itemize}

\section{Related work}
\label{sec:related_works}
\sloppy Machine reading comprehension (MRC) and Open Domain Question Answering are two problems that are pursued actively by the NLP and IR communities.
Introduction of large scale datasets like the Stanford Question Answering Dataset (SQuAD1.0)~\cite{squad}, MicroSoft MAchine Reading COmprehension dataset (MS MACRO)~\cite{ms_marco}  and Natural Questions~\cite{natural_questions} have led to the development of deep learning based QA\slash MRC systems~\cite{hermann2015teaching,shen2017reasonet,wang2016machine,bert} that can answer questions about a given a corpus of text or passage. 
The main difference between these works with the problem we address here is that they use textual information provided as computer readable strings (electronic text), while we tackle the problem of answering questions asked on a set of document images without any given transcription. 

Our work is also related to the problem of VQA, which is receiving increasing interest from the Computer Vision research community in recent years~\cite{antol2015vqa,gao2015you,ren2015exploring,goyal2017making,johnson2017clevr,agrawal2018don}.
Most of the early VQA datasets and methods disregard text present in the images, and the problem is often modelled as a multi-class classification where the set of output answers is fixed.
Questions in these datasets typically focus on visual aspects such as objects, attributes, and relationships.

Gurari et~al.~\cite{vizwiz_challenge} showed that in a goal oriented VQA setting where visually impaired individuals ask questions on images they take, answering a good number of questions require the ability to read and interpret text on the images. This inspired introduction of two datasets --- Scene Text VQA~\cite{stvqa} and TextVQA~\cite{textvqa}, where reading text on the images is pivotal to answering the questions asked on the images.
Our work is different from these tasks on two accounts: (i) these datasets contain images ``in the wild'' which are drawn from popular scene text datasets or datasets like OpenImages~\cite{openImages2}, which predominantly have scattered text tokens compared to the handwritten document images we consider, and (ii) almost all VQA problems including the ones involving text on the images are formulated as QA on a single image, while the proposed QA task is for a collection of document images.

Another set of VQA works related to ours is VQA on charts and plots ~\cite{kafle2018dvqa,kahou2017figureqa} 
and the work of Kembhavi et~al.~\cite{kembhavi2017you} on Textbook Question Answering (TQA).
The TQA dataset aims at answering questions given a context of text, diagrams and images. However textual information is provided in computer readable format.
But in case of the VQA on charts and plots, text on the images needs to be recognized using an OCR to answer a good number of questions.
However, the text found on the synthetically generated charts/plots on these datasets is sparse and rendered in standard font types and in good quality, compared to the handwritten text in the form of sentences and paragraphs (running text) in our case.

As stated earlier, this work is motivated by our own recent work on VQA on document images called DocVQA.
The dataset for task 1 of DocVQA 
comprises a wide variety of documents containing printed, typewritten, handwritten and born-digital text. The documents include content in the form of sentences, forms, tables, figures and photographs.
While DocVQA task 1 sticks to the standard VQA setting where the answer is textual, we propose to respond to natural language queries by providing ``visual answers'' in the form of image snippets.

On the information retrieval and keyword spotting front,
there are a plethora of works dealing with
handwritten document indexing and retrieval~\cite{hw_retrieval_anoop,hw_retrieval_howe,hw_retrieval_Cao,hw_retrieval_ahmed2017survey,imageclef}.
One relevant example is the ImageCLEF 2016 Handwritten Scanned Document Retrieval challenge~\cite{imageclef}, aimed at developing retrieval systems for handwritten documents. 
Although there are some similarities between the ImageCLEF 2016 document retrieval challenge and the QA on handwritten documents proposed in this paper---their queries have multiple words (like our questions) and the retrieval instance is a document segment/snippet---the task of document retrieval which they address differs clearly from  the proposed QA in the following aspects: (i) queries in their case are not natural language questions but search queries having multiple tokens and (ii) their task requires that all the tokens in the input query appear in the same order in the retrieved document snippet.

Kise et al.~\cite{kise} address the problem of document retrieval for building a QA system for a collection of printed document images.
This probably is the first work on QA over a document image collection.
In their work they use documents  with machine printed English text. Recognition of printed text is relatively easier and  allows use of recognition-based systems for QA over such collections.
Hence, contrary to the recognition-free approach proposed in our work,  their approach first recognizes the text from the documents and the transcribed electronic text is the input to their QA system.
Task 2 of DocVQA~\cite{docvqa_challenge} is similar to the work we present here for the fact that both deal with the problem of QA over a document collection. The former uses a document collection in which all documents are forms of the same template (US candidate registration forms), compared to the handwritten documents in HW-SQuAD and BenthamQA with diverse content. The dataset has 20 questions in total.
Unlike the proposed formulation to return answer snippets, the objective is to retrieve all documents (or forms) in the collection required to answer the question correctly. 

\begin{figure*}[htp]
\centering
\includegraphics[width=\linewidth]{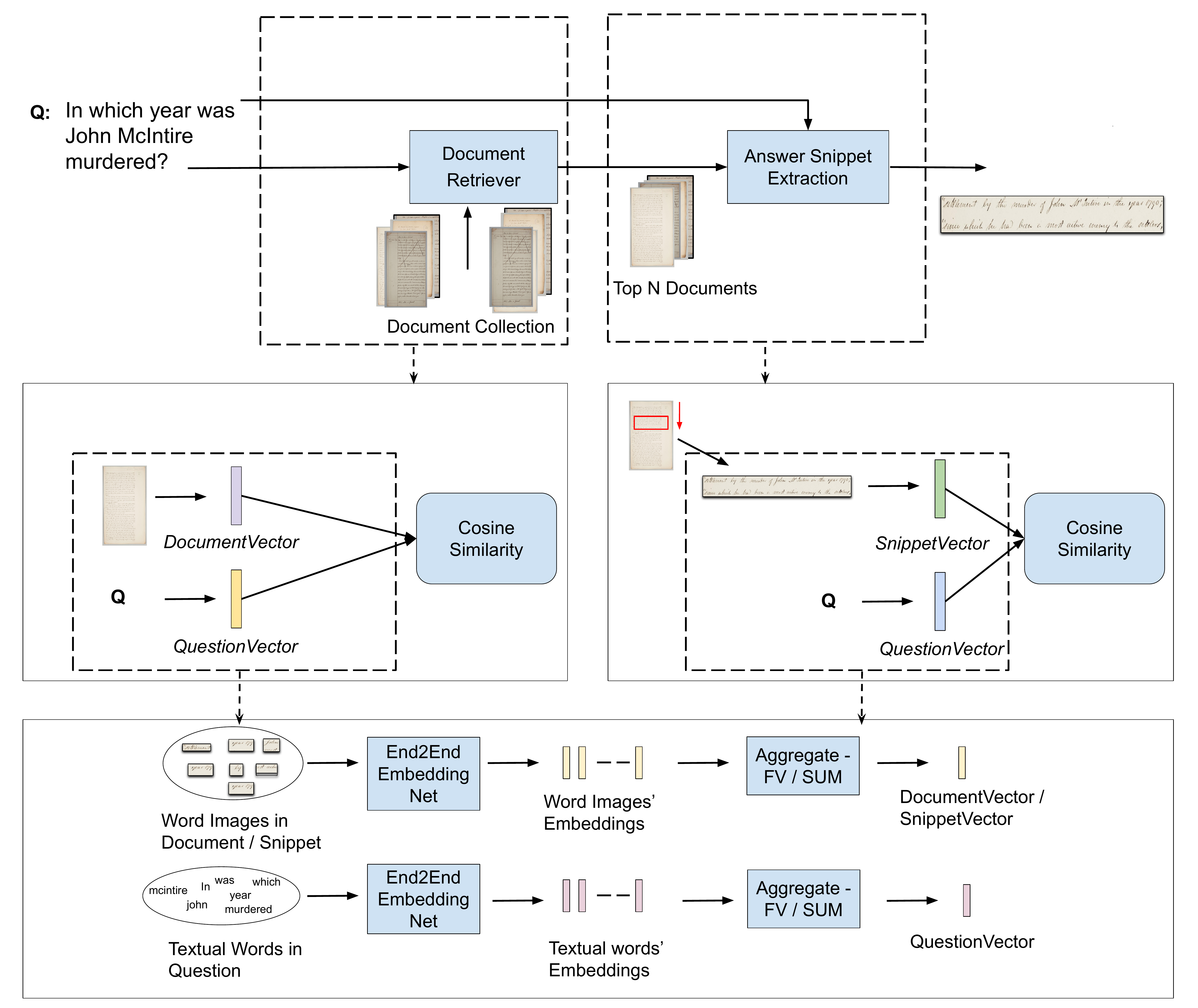}
\caption{Proposed recognition-free approach to QA on document collection. Our approach works by aggregating embeddings of words in a question, a handwritten document or a document snippet to a $QuestionVector$, $DocumentVector$ or $SnippetVector$ respectively. We use an end-to-end embedding network~\cite{praveen_e2e} for the embeddings. These vector representations help us to retrieve documents similar to a question or snippets similar to a question in a Vector Space model. The snippets are extracted in a two step process. In the first step, a small number of relevant documents are retrieved. In the second step, a document snippet answering the question is extracted from one of the previously retrieved documents.}
\label{fig:model_architecture}

\end{figure*}

\section{Method}
\label{sec:section3_method}
In the following discussion, we use the term `document' to refer to a handwritten document image. 
`Word' refers to a textual word in questions and a word image in documents.
We define a document snippet or  a `snippet' as a horizontal slice of one or more contiguous text lines from a document.

The proposed QA method returns the answer to a given question as a snippet extracted from one of the queried documents.
Given a question like \textit{In which year was John McIntire murdered?} (see Fig.~\ref{fig:fig1:the_process}) we would like to find an embedding\slash representation for the question.
We aim to find a snippet in the new embedding space that can be the most probable answer using a simple nearest neighbour search.
This approach leads to the following technical challenges:
\begin{itemize}
    \item We need an embedding where both text (questions) and image (document/snippet) modalities can be embedded.
    \item The embedding of a question must be independent of its length and its paraphrasing.
    \item In the joint embedding space, a question's matching snippet must have its embedding similar to the question.
\end{itemize}

We describe a solution that meets the above requirements below. Fig.~\ref{fig:model_architecture} shows a schematic representation of the proposed solution.

\subsection{Joint embedding of text strings and word images}
\label{method_word_embeddings}
In order to come up with a joint space where questions and snippets can match, we first embed the constituent words of a question or snippet to a joint space.
Later these  individual embeddings  are aggregated to form a single global embedding for the question or the snippet.
For the joint embedding of words, we use an end-to-end embedding network introduced in~\cite{praveen_e2e}, which simultaneously learns both text and image embeddings using a multi-task loss.
The primary advantage of this model is that it is an end-to-end trainable network, without the need to embed image and text in separate steps and later combine them.. 
The network learns a feature space where words which are 
lexically similar lie closer to each other. This facilitates word spotting in this joint space, in both query-by-string and query-by-example settings. Word spotting results using this network, suggest that it does a  better job at learning joint representations compared to other similar approaches in the recent years~\cite{phocnet,tppphoc,lsde}.
\subsection{Aggregating word embeddings}
\label{method_aggregation}

The answer extraction process we discuss below requires us to represent questions, documents, and snippets as vectors of a fixed dimension.
This is achieved by aggregating the embeddings of the constituent words of a question\slash document\slash snippet into a global representation, whose size is independent of the number of words being aggregated.
Stop words in questions and documents\slash snippets are removed prior to the aggregation of word embeddings (see Section~\ref{subsec_implementation_details} for details). 
Two different approaches are tried out for aggregating the word embeddings and they are discussed below.
\subsubsection{Aggregate by Summing (SUM)}
\label{agg_sum}
A set of $M$ word embeddings $\{x_1, x_2....x_M\}$ of size $D_w$ are summed together to form an aggregate vector of the same size.
\subsubsection{Aggregate using Fisher Vector (FV) framework}
\label{agg_fisher}
The Fisher Vector (FV) framework introduced in~\cite{Jaakkola} has been widely used to aggregate local image descriptors to a global descriptor in image retrieval and classification problems~\cite{fisher_original,fisher_inria,fisher_improving}. Given a set of word embeddings $X = \{x_1, x_2....x_M\}$ we assume that these continuous word embeddings have been generated by
a Gaussian Mixture Model (GMM).
We denote the parameters of the GMM, $\lambda=\{{w_i},{\mu_i},{\Sigma_i},\:i = 1\ldots K\}$, where $K$ is the number of Gaussians. $w_i$, $\mu_i$ and $\Sigma_i$ are respectively the mixture weight, mean vector and covariance matrix of the Gaussian $i$.
We assume that the covariance matrices are diagonal and hence it is denoted by the variance vector, $\sigma^2_i$.
The GMM, $u_\lambda$ is estimated/trained offline from a representative set of samples using Maximum Likelihood (ML) estimation.

For the $i^{th}$ Gaussian in the GMM, let $\mathcal{G}^\lambda_{\mu,i}$ be the gradient with respect to the mean ($\mu_i$) and $\mathcal{G}^\lambda_{\sigma,i}$ be the gradient with respect to the standard deviation ($\sigma_i$).
FV of the given set of embeddings $X$ will be the gradient vector $\mathcal{G}^\lambda_{X}$ which is obtained by concatenating $\mathcal{G}^\lambda_{\mu,i}$ and $\mathcal{G}^\lambda_{\sigma,i}$ for $i=1 \ldots K$.
Following the standard practice we do not consider gradients with respect to the mixture weights, $w_i$s, since it adds little discriminative information~\cite{aggregate_fisher_inria,fisher_improving}.
Finally, we apply Power normalization and L2 normalization to the FV as suggested in~\cite{fisher_inria}.
Power normalization proposed  by Perronnin et al.~\cite{fisher_improving} helps to ``unsparisfy'' the FV as it becomes sparser with more number of Gaussians. To each dimension of the FV, power normalization applies the following function $f(z)=sign(z){|z|}^\alpha$. $0 \leq \alpha \leq 1$ is the power normalization parameter and usually set to $1/2$.

\subsection{Document retriever}
\label{method_doc_retreivl}
Following the standard practice in QA systems for electronic text~\cite{reading_wiki_acl}, we first select a small set of documents that are likely to contain the answer. We call this step ``Document Retriever''.
Given a question and a document collection of $M$ documents, the Document Retriever generates $N$ proposals ( $N << M $ ) for the `target document', i.e., the document where the potential answer to a given question lies. 
For this step, the question and the documents are summarized into a $QuestionVector$ and $DocumentVectors$ of fixed size using one of the aggregation methods mentioned above.
\sloppy Subsequently, target document proposals are generated by retrieving $N$ documents best matching the question by computing cosine similarity between the $QuestionVector$ and  the $DocumentVector$s. 

\subsection{Answer snippet extraction}
\label{method_snippet_retreival}
Once a set of target document proposals are generated by the Document Retriever, all possible snippets from these proposals are considered as `candidate snippets'. Since our approach assumes that word and line segmentations are known, the candidates are extracted in a sliding window manner, scanning text lines in a document from top to bottom. For instance if the size of answer snippets is set as 2 text lines, we group each pair of consecutive text lines as a snippet. 
First two lines in a document as one snippet, second and third lines as another one and likewise the second last and last lines will form the last candidate snippet.
Then, candidate snippets are converted into $SnippetVectors$ using one of the aggregation methods proposed in Section~\ref{method_aggregation}.
The aggregation method used for this step could be different from the one used with the document retriever.
The best aggregation scheme for each step shall be found empirically, by evaluating different options on a validation set. Finally, similar to Document Retriever, cosine similarity between the $QuestionVector$ and $SnippetVectors$ are computed and the top matching snippet is returned as the answer to the given question.

\begin{figure*}[htp]
\centering
\includegraphics[width=\linewidth,height=5cm]{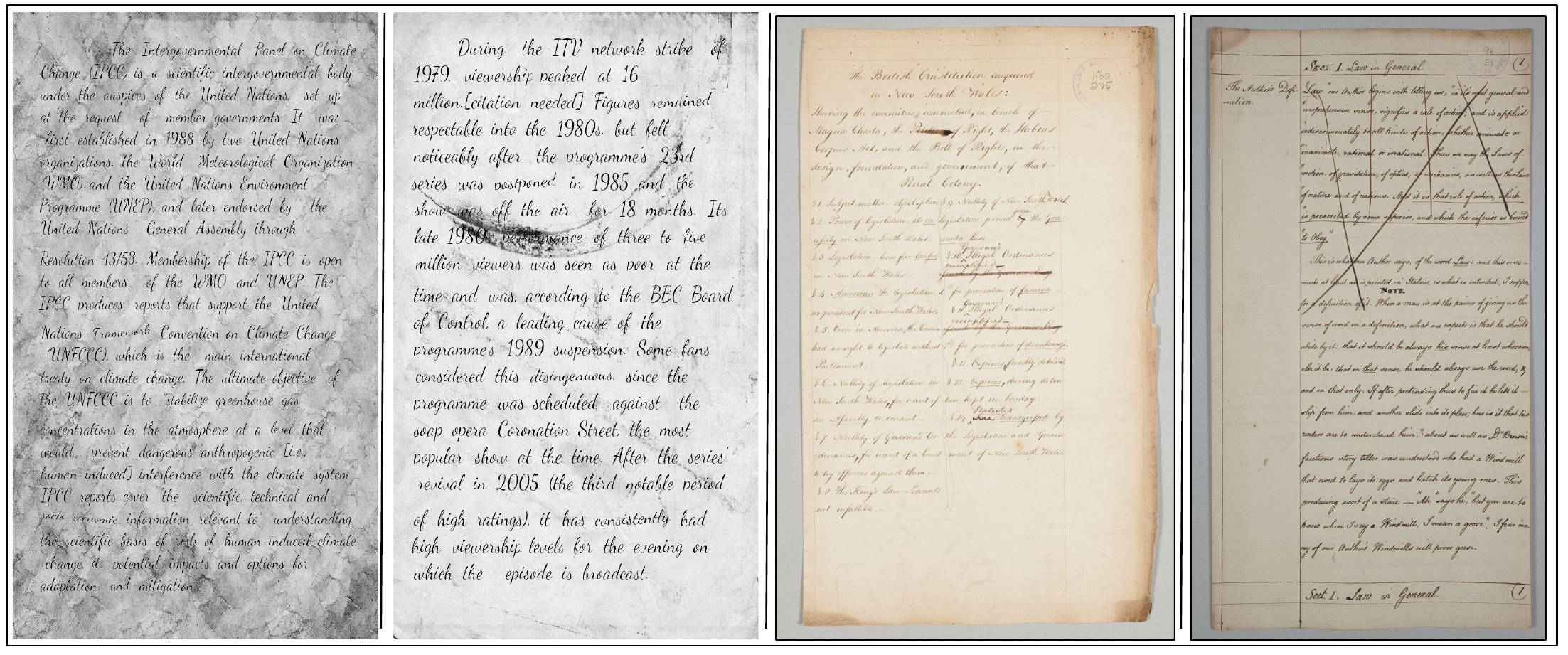}
\caption{Sample documents in the newly introduced HW-SQuAD and BenthamQA datasets. The first two are synthetically rendered images in HW-SQuAD.
The synthetic samples have diverse fonts, backgrounds, textures, image artifacts and inter-word, and inter-line spacing.
The last two are real handwritten manuscripts which are part of the BenthamQA dataset.}
\label{fig:document_examples}

\end{figure*}
\section{HW-SQuAD and BenthamQA datasets}
\label{section_hw_squad_dataset}
The annotation of a QA dataset on a handwritten document collection requires considerable human effort.
Moreover, modern deep learning-based supervised models benefit from large amounts of  training data. As an alternative to costly human annotation, we build a dataset using an existing QA dataset for electronic text by reusing the questions and answers.
Since this dataset is a derivative of the existing SQuAD1.0 dataset, we term it HW-SQuAD. 
We use HW-SQuAD for experimenting with different ablations of the proposed approach. For benchmarking the approach on a real collection of handwritten manuscripts we create another dataset called BenthamQA.
Images in BenthamQA are scanned images of handwritten manuscripts from 18\textsuperscript{th} and 19\textsuperscript{th} centuries.

\subsection{HW-SQuAD}
\label{subsec:hwsquad}
SQuAD1.0~\cite{squad} is a popular benchmark dataset used by NLP and IR communities for QA and MRC on electronic text. It contains 100,000+ question--answer pairs on 23,000+ passages, drawn from over 500 Wikipedia articles.
Every question is associated with a passage, and the answer to the question is marked as a `span' of contiguous words in the passage.
The train set of the dataset has only one answer mapped per question. But in the development (public test) set, each question is provided with one or more target (ground truth) answers.
The analysis presented in~\cite{squad} shows the dataset has diverse questions in terms of  i) syntactic and lexical variations, and ii) the extent of reasoning required to answer the questions.

The passages need to be rendered as document images to reuse SQuAD1.0 for our purpose.
Motivated by the success of synthetic datasets for OCR problems~\cite{jaderberg,praveen_e2e}, we synthetically render the passages in SQuAD1.0 as handwritten document images. In the following, we describe how we use SQuAD1.0 data to create the new HW-SQuAD dataset.

\subsubsection{Data preprocessing}
\label{subsubsec:content_curation}
Samples in SQuAD1.0 are first subjected to standard pre-processing steps\footnote{\url{https://github.com/facebookresearch/DrQA}} to tokenize questions and passages, and to generate question--passage correspondences in terms of token offsets.
Non-ASCII tokens in questions passages are then converted to nearest ASCII transliterations using the \texttt{Unidecode} library\footnote{\url{https://pypi.org/project/Unidecode/}}. This was required since most of the handwritten fonts we use do not support characters outside the ASCII character set.
Following the pre-processing step, a few questions---corresponding to the answers for which mapping the character offsets to the token offsets fail---are excluded.
The original train set of SQuAD1.0 is split randomly into two subsets in HW-SQuAD --- 95\% in train and 5\% for validation (val). We use the original development set (public test set) of SQuAD1.0 to create the test set in HW-SQuAD.

\begin{table*}[tp]
\begin{center}
\begin{tabularx}{0.95\linewidth}{X c c c c}
& \multicolumn{3}{c}{HW-SQuAD} & BenthamQA \\
&train & val & test &  \\ \toprule
Number of Questions & 67887 & 7578 & 9477 & 200\\
Number of Documents & 17007 & 1889 & 2067 & 338 \\
Average number of words per document & 116.6 & 117.2 & 122.8 & 202.0 \\ 
Average number of words per question & 10.1 & 10.1 & 10.2 & 18.0\\
Average number of words per answer & 2.4 & 2.4 & 2.7 & 3.9\\
\bottomrule
\end{tabularx}
\end{center}
\caption{HW-SQuAD and BenthamQA dataset statistics.}
\label{table_squad}

\end{table*}

\subsubsection{Synthetic rendering of SQuAD1.0 passages as handwritten document images}
\label{subsubsec:synth_rendering}
We render a passage as a single-column document image using a handwriting font. The font used is randomly sampled from over 100 handwriting style fonts downloaded from Google Fonts\footnote{\url{https://fonts.google.com/}}.

The first step is to render each word in a passage in SQuAD1.0 onto a transparent background.
The font size is set randomly from a range of 28--52 pts. 
Intensity of the foreground ( i.e., the pixels making up the text strokes) is varied randomly in the 0--50 range (0 being darkest and 255 being the brightest). This means all our documents have darker text on lighter backgrounds, which is the case for  document images in general. For a random 15\% of the word images, we induced degradation by eroding the images using the \texttt{erode()} function available in OpenCV~\cite{opencv_library}.

In the next step, all the word images are pasted onto a background image in the same order as in the original passage using alpha blending. The background image is randomly sampled from a set of 20+ manuscripts like textures downloaded from the internet.
While pasting the words onto the background image, we break the lines after every few words.
The number of words in a line is varied randomly, with a minimum of 5 and a maximum of 7. 
For a document, the fixed inter-word spacing used is the average per character width (width of a word image divided by number of characters in it) for the words in the document. Similarly the fixed inter-line spacing  used for a document is the average height of word images in the document. For a random 15\%, we used a different spacing than the fixed value. In such cases, we multiply the fixed value by a random multiplier in the range of 0.9--2.5 for inter-word spacing and a random multiplier in the range of 0.9--1.3 for inter-line spacing. The borders on all 4 sides of the document are also set as a function of the inter-word spacing used in the document. The border on each side is the fixed inter-word spacing for the document multiplied by a random value in the range 1.5--5.

Once all word images are pasted onto the background image, we
subjected the image to a random amount of skew. The skew angle we used is a random value in the range -5--5. Then a random 15\% of the images are re-sampled and brought back to the original dimensions to induce artifacts resulting from re-sampling of images. For re-sampling, we used a random scaling factor in 0.6--1.4 range.

\subsection{BenthamQA}
\label{subsec:BenthamQA}
To build a handwritten QA dataset containing real images, we use manuscripts from historical collections since OCR is typically poorer on such images. We use a collection with word and line bounding box annotation already available so that QA methods that require gold-standard annotations for words or lines can use them. We use the ImageCLEF 2016 Bentham Handwritten Retrieval dataset~\cite{imageclef}, which has images from the Bentham Transcriptorium project~\cite{bentham_transcribe}. 433 images in the development (dev) split of this dataset have manually reviewed bounding box annotations for words and lines. We chose this split as the document collection for our QA annotation.
Comparing the documents using Jaccard Similarity of their gold-standard transcriptions, duplicates and near-duplicates were filtered out. 
In HW-SQuAD, we could define answers in the image space since we know the words that constitute the answer and their bounding boxes. Therefore, to annotate the Bentham manuscripts, we used a tool that allows annotators to see the image, add questions and textual answers, and mark the region where the answer lies. During the pilot annotation, the annotators had difficulty reading text on the manuscripts, thereby slowing down the annotation process. Hence we tested an approach similar to the annotation of SQuAD-like datasets.
We used a hosted version of the Haystack annotation tool\footnote{\url{https://annotate.deepset.ai}} for the same.
The gold-standard transcriptions of the manuscripts were shown to the annotators to add questions and answers. Similar to the typical extractive QA annotation, the answers were marked as spans of the transcripts. The textual words that constitute each answer were then mapped to the corresponding image locations. With the help of two volunteers, 200 question--answer pairs based on 94 different document images were collected.
However there are total 383 images in the collection.
We keep the remaining images in the dataset as distractors for the QA task. In other words, we have 200 questions on 338 images and  94 images have at least one question based on it.

Basic statistics of questions, answers and documents in  both HW-SQuAD and BenthamQA are shown in Table~\ref{table_squad}. 
Questions in BenthamQA are almost double the length of questions in HW-SQuAD. This is expected since SQuAD1.0 dataset was annotated for MRC where each question is asked in the context of a single passage. A detailed discussion on this aspect is given in Section 5.3 when we analyze the performance based on length of questions.
While   annotating BenthamQA, annotators were instructed not to ask questions specific to a particular document such as  \textit{What is the title of the document?} or \textit{What is the name of the city mentioned in the first paragraph?}.
This made the annotators ask longer questions. For example, take this  question in the dataset, \textit{In one of the notes where the author talks about ``Genus" and ``Species" in the context of law, he makes a comparison with a flower. Which flower is it?'}. The question gives context that helps find the target manuscript from the collection and then talks about the context within the  manuscript.
On average, document images in BenthamQA have more words in them 
since these documents are manuscript pages compared to passages rendered as document images in HW-SQuAD. 
We show two documents each from HW-SQuAD and BenthamQA in Fig.~\ref{fig:document_examples}.

\subsection{Performance evaluation}
\label{subsec:evaluation}

QA and  MRC benchmarks  use evaluation metrics like F1 score and Exact Match~\cite{squad,reading_wiki_acl}. On the other hand, standard VQA benchmarks, particularly those that do not involve text, use a metric called VQA Accuracy. It treats a predicted answer as correct if at least 3 of the 10 annotators  entered the exact same answer~\cite{antol2015vqa}. Scene Text VQA and DocVQA use a metric, named Average Normalized Levenshtein Similarity (ANLS)~\cite{stvqa}. The metric assigns a score for each prediction based on the Levenshtein distance between the prediction and the ground truth answer.
Since all these metrics are conceptualized for textual answers, they cannot be used for a QA task where answers are document snippets.
Hence  new evaluation scheme is devised where the predicted answer snippet is evaluated  against a ground truth defined in the image space.

\begin{figure}[tp]
\setlength{\tabcolsep}{2pt}
\renewcommand{\arraystretch}{0.25}
\frame{\begin{tabularx}{\linewidth}{l m{0.4cm}l}
&&\\
&&\\
&\small{\fontfamily{qhv}\selectfont } & \small{\fontfamily{qhv}\selectfont Telnet used what interface technology? }\\
&&\\
&&\\
&{\small{\fontfamily{qhv}\selectfont \color{red}A1:}} & ``host interface to X.25 and terminal interface to X.29" \\
&&\\
&&\\
&{\small{\fontfamily{qhv}\selectfont \color{blue}A2:}} & ``X.25" \\
&&\\
&&\\
&{\small{\fontfamily{qhv}\selectfont \color{green}A3:}} & ``ARPANET" \\
&&\\
&&\\
\multicolumn{3}{c} { \includegraphics[width=0.9\linewidth,height=8.5cm]{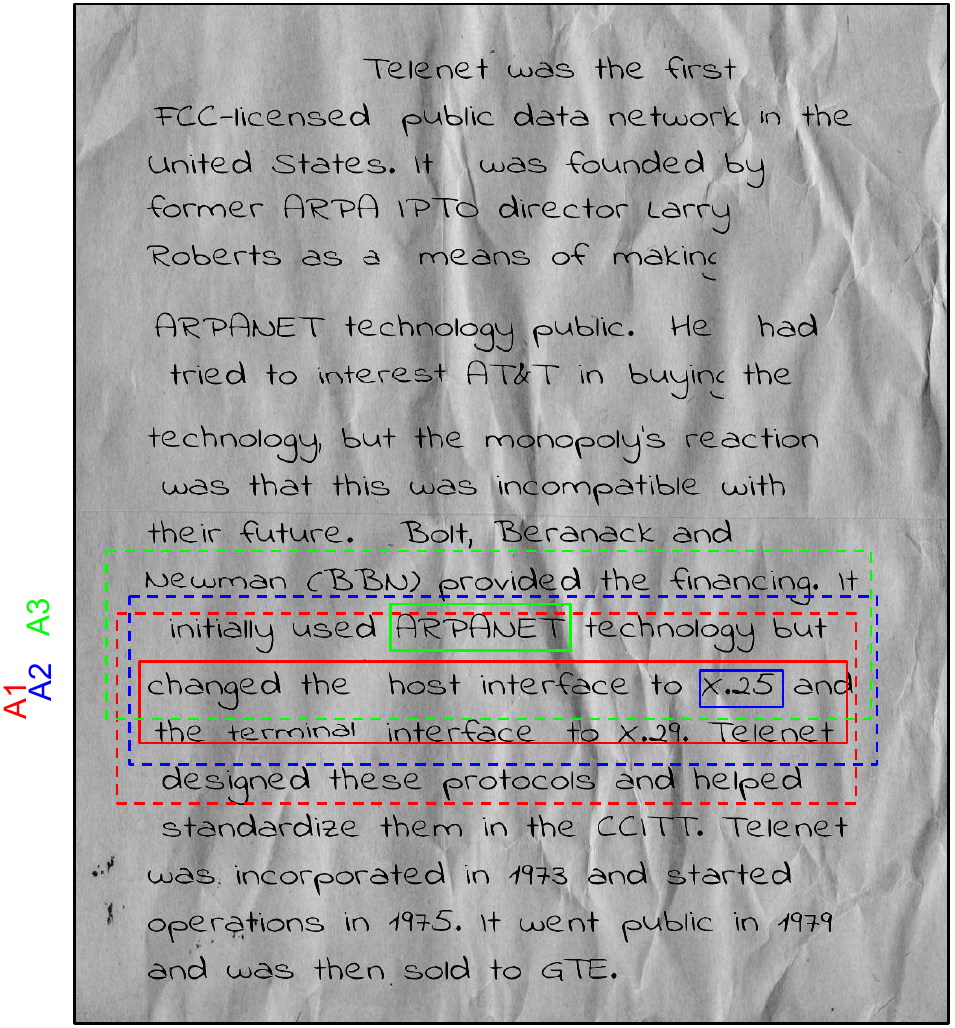} } \\
&&\\
&&\\
\end{tabularx}}
\caption{A question--answer pair and the associated document from the test set of HW-SQuAD dataset. To evaluate answer snippet extraction, we define ground truth in the image space in terms of a Large Box ($LB$) which includes the lines where the answer is, plus a line above and below it and a Small Box ($SB$) which tightly encloses the words constituting the answer. In this case there are three ground truth textual answers, and hence we show $SB$ (solid lines) and $LB$ (dotted lines) for the three answers. Note: the margins of the boxes shown in the figure is changed a bit to show the overlapping boxes without clutter.}
\label{fig:hw_squad_annotation}

\end{figure}

Wang et al.~\cite{general_value_of_evidence}
proposes a VQA task where models need to predict a textual answer and a bounding box as `evidence'. The ground truth `evidence' is used to assess the predicted `evidence'. Inspired by evaluation in object detection benchmarks, the authors propose to use Intersection Over Union (IOU) as a measure to evaluate the predicted box.
We considered a similar evaluation scheme, by treating the minimum rectangle bounding the answer words as the ground truth (GT) box.
Methods which work without recognizing the text in the documents such as the method we propose in this work, do not understand the semantics of the text and must be relying on similarity between words in question and the context where the answer lies.
It is unlikely that such approaches can narrow down to a tight box around the answer words. This would result in low IOU scores when pitted against a GT box which tightly bounds the answer.
An alternative is to check if the GT box is inside the predicted snippet. 
However a method can always output bigger snippets that contain the answer words and satisfy the condition.
This motivated us to develop an evaluation scheme, which accommodates methods that return answers as short snippets containing the exact answer and some context.
At the same time we want to make sure that larger snippets with too much context are discouraged.
\\
We treat the predicted answer snippet (or Answer Box; $AB$ for short) as a correct prediction if: (i) $AB$ is inside a Large Box ($LB$) which is the region in the image enclosing the text lines where the words making up the ground truth answer lie and one line above and one line below it and (ii) the smallest box containing the words constituting the textual answer (i.e., a Small Box or $SB$) is inside the $AB$. In short, the condition for an answer snippet to be a correct prediction is a double inclusion criterion: $AB \subset LB \land SB \subset AB$. To understand this better we show $LB$ and $SB$ for an example case in Fig.~\ref{fig:hw_squad_annotation}. 
To incorporate the double inclusion criterion into a score, we define Double Inclusion Score (DIS) as

\begin{equation*}
    DIS = \frac{{AB} \cap {SB}}{|SB|} \times  \frac{{AB} \cap {LB}}{|AB|}
\end{equation*}

For a question, if the predicted answer snippet has a $DIS > 0.8$ with any of the ground truth answers, we consider it as a correct prediction. Accuracy is then calculated as the percentage of questions for which we have a correct prediction.
Note that the HW-SQuAD and BenthamQA datasets can also be used for recognition-based QA where textual answers are expected. In such a setting, it is ideal to use the standard metrics  for text based QA, such as Exact Match (EM) score and F1 score~\cite{squad}. In Section~\ref{subsec:recognition_based}, we report results on both the datasets in recognition-based setting where textual answers are expected.

\section{Experiments and results}
In this section we present experimental settings, results of our experiments, and a discussion of the results.
\subsection{Implementation details}
\label{subsec_implementation_details}
Questions are split into constituent words, and stop words are removed using NLTK~\cite{nltk}.
Words from the documents are extracted using gold-standard word bounding boxes available with the datasets. In practice, word bounding boxes are obtained using a document segmentation algorithm, and the segmentations need not be perfect. 
Hence results we report below do not factor in possible segmentation errors in such scenarios. To filter out stop words from the documents, we train a Convolutional Neural Network (CNN) based binary classifier. The classifier separates word images from the documents into stop words and non-stop words.
The feature extraction block of this network follows the same architecture as the one in CRNN~\cite{crnn} network. Feature maps after the last convolutional layer are mapped to a binary classification layer.
The classifier is trained using synthetic, handwritten word images in HW-SYNTH dataset~\cite{synth-data} and word images from IAM dataset~\cite{iam}.

After stop words removal, words from both the questions and documents are fed to the end-to-end embedding network to represent them as word embeddings.
We use a pretrained model\footnote{\url{https://github.com/kris314/e2eEmbed}} for this network, made publicly available by the authors. The pretrained model is trained on HW-SYNTH and IAM datasets. The output of the network is an L2 normalized embedding vector $x \in \R^{2048}$. We use Principal Component Analysis (PCA)~\cite{bishop} to map $x$
to a $x^{\prime} \in \R^{Dw}$ in a lower dimensional space. PCA rotation matrices are learnt on the training set of the HW-SQuAD.
In our experiments, particularly while using FV aggregation scheme, we work with these dimensionality reduced vectors.

Parameters of the GMMs are estimated offline using embeddings of around $1.2$ million words (after filtering stop words) drawn from documents in the train set of HW-SQuAD. Since words in questions are almost a subset of the words in the documents, and word images add more variability than textual words, we use only words from documents to fit the GMMs.
To get candidate snippets from the document proposals generated by the document retriever (see Section~\ref{method_snippet_retreival}), we use sliding windows of size $2$ and a step size of $1$ over the text lines, scanning from top to bottom. For example, from a document proposal having $l$ text lines, we have $l-1$ candidate snippets, each one enclosing two contiguous text lines.

 \begin{table}[t]
 \begin{center}
    
 \begin{tabularx}{\linewidth}{X  c c c}
& \multicolumn{2}{c}{HW-SQuAD} & BenthamQA \\
$D_w$ & val & test  &   \\ \toprule 

  2048 & 26.0 & 22.1 & 23.5 \\
 200 & 21.8 &  & \\
 128 & 20.5 &  & \\ 
  64 &  16.3 &  & \\ \bottomrule 
 \end{tabularx}
 \end{center}

\caption{Top-5(\%) accuracy for the document retriever when the aggregation scheme used is SUM.}
 \label{table_doc_sum}

\end{table}

\subsection{Performance of document retriever}
\label{exp_doc_proposal}
.

\begin{table*}[t]
\small
\begin{center}
\begin{tabularx}{\linewidth}{X ccccccc}
& \multicolumn{6}{c}{HW-SQuAD} & BenthamQA \\
  &\multicolumn{5}{c}{val} & test &  \\ \toprule
\diagbox{$D_w$}{$K$} & 32 & 64 & 128 & 256 & 512 & 512 & 512 \\ \midrule 
\multicolumn{7}{c}{ \textit{Gradients w.r.t to both mean and standard deviation considered i.e. } $FV$ = $\mathcal{G}^\lambda_{\mu,i}$s and $\mathcal{G}^\lambda_{\sigma,i}$s ; ${|FV|} = 2K{D_w}$} \\ \midrule 
 64 & 14.6 & 15.6 & 16.2 &   \\
128 & 10.8 & 14.4 & 17.5 &   \\
200 & 6.8 & 12.9 & 15.4 &   \\ \midrule 
\multicolumn{7}{c}{\textit{ Only gradients w.r.t to mean are considered i.e} - $FV$ = $\mathcal{G}^\lambda_{\mu,i}$s ;  ${|FV|} = K{D_w}$} \\ \midrule 
  64 & 31.9 & 39.5 & 40.9 & 44.0 & 48.2 &  \\
 128 & 36.6 & 43.2 & 45.7 & 48.3 & \textbf{50.4}  & \textbf{46.5} & \textbf{55.5} \\
 200 & 30.1 & 40.9 & 42.0 & 44.7 & 46.5 & \\  \bottomrule 
 
\end{tabularx}
\end{center}

\caption{Top-5 accuracy (\%) for the document retriever when FV aggregation is used.
The top half shows the results when FV is a concatenation of gradients with respect to means ($\mathcal{G}^\lambda_{\mu,i}$) and standard deviations of the Gaussians. ($\mathcal{G}^\lambda_{\sigma,i}$). The bottom half shows results when FV formed is a concatenation of gradients with respect to means alone. In all cases the resulting FVs are normalized using     power normalization and L2 normalization.}
\label{table_doc_fisher_full}

\end{table*}

Since the performance of the answer snippet extraction (Section ~\ref{method_snippet_retreival}) step is dependent on the quality of the document proposals generated by the document retriever (Section~\ref{method_doc_retreivl}), we first evaluate the document retriever. Top-$N$ accuracy---percentage of questions for which target document is in top $N$ proposals---is used as the evaluation metric. Table~\ref{table_doc_sum} reports top-$5$ accuracy for the proposals retrieval on both the datasets for different word embedding sizes ($D_w$) when the aggregation scheme used is SUM. We appreciate that the original embedding vector (without any dimensionality reduction) of size $2048$ yields the best results in this case.

\begin{figure}[tp]
    \centering
    \includegraphics[width=0.85\linewidth]{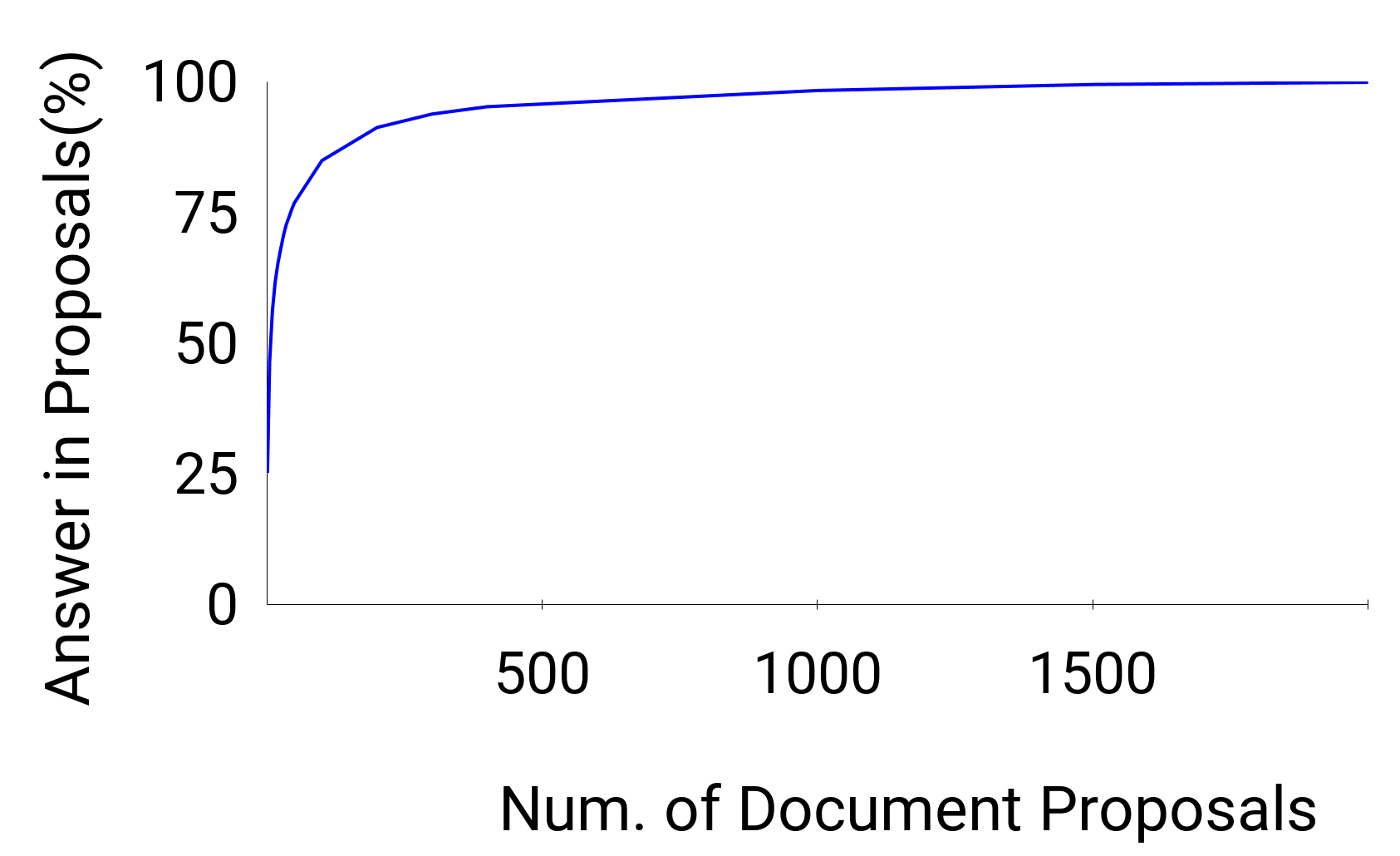}
    \caption{With more %
   document proposals generated for the target document, the chance of finding the answer to a question in the document proposals increases. When the number of document proposals is more than 2000, the answer is found within the proposals for almost all the questions. The given plot is for validation set of HW-SQuAD.}
    \label{fig:propoasl_vs_accuracy}
\end{figure}

Performance of the document retriever using FV aggregation for different word embedding sizes and number of Gaussians in GMM ($K$) is shown in Table~\ref{table_doc_fisher_full}.
The top half of the table shows results when the FV is a concatenation of gradients with respect to both the mean ($\mathcal{G}^\lambda_{\mu,i}$) and standard deviation ($\mathcal{G}^\lambda_{\sigma,i}$) of the Gaussians. Thus the size of the resulting FV ($\lvert FV \rvert$) is $2KD_w$, since each gradient vector is of size $D_w$ (see section~\ref{agg_fisher}). 
We repeat the same set of experiments using FVs where only the gradients with respect to the means of the Gaussians are concatenated to form the FV. These results are shown in the bottom half of the table.
This setting yields much better performance, consistent across different values of $K$ and $D_w$.
It also helps in faster computations since $\lvert FV \rvert$ becomes half in this case. For this reason, experiments considering gradients with respect to both the mean and standard deviation of Gaussians are not conducted for higher values of $K$.
On the test set of the HW-SQuAD and on BenthamQA, we report results only for the setting which yields the best performance on the validation (val) set of HW-SQuAD.

The best performance on the validation set is for $K=512$ and dimensionality reduced word embeddings of size $D_{w}=128$.
Embeddings of size 128 give better results compared to a larger embedding size of 200 for all values of $K$.
This is in line with the finding in~\cite{aggregate_fisher_inria} that dimensionality reduction of the local features using PCA improves the overall performance of classification and retrieval. 
In \cite{aggregate_fisher_inria} the authors cite two possible reasons for this: (i) decorrelated data can be fitted more accurately when the covariance matrices of the GMM are diagonal, and (ii) GMM estimation is less noisy if only the `stronger' components are considered.
In Table~\ref{table_doc_fisher_full}, it can be seen that the larger the number of Gaussians $K$, the better the retrieval performance. However, for higher values of $K$ dimensionality of the FV becomes very large and it is computationally expensive to work with such large vectors in practical scenarios.

\begin{figure}[]
\centering
\includegraphics[width=\linewidth]{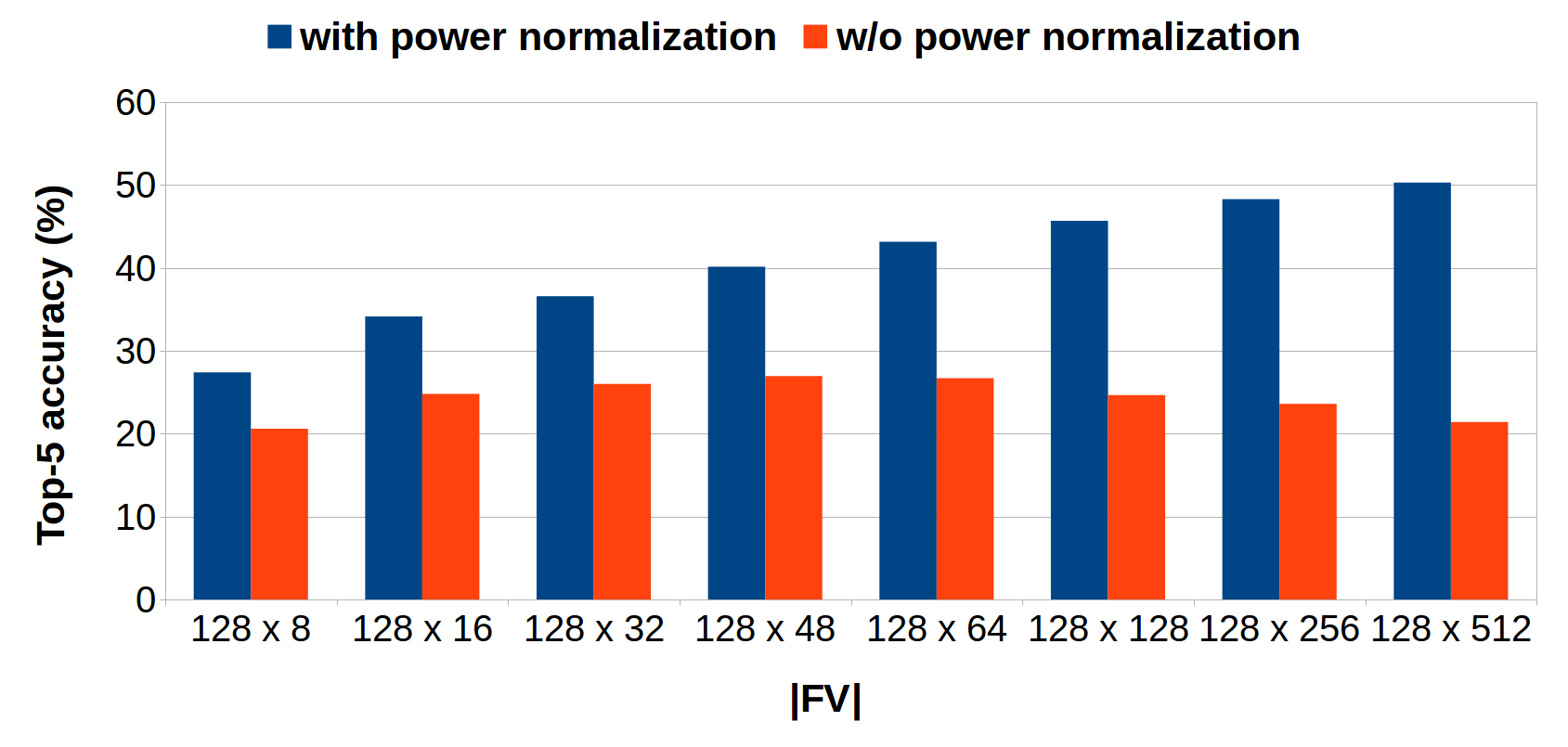}
\caption{ Top-5 accuracy for document retriever for different sizes of $FV$, with and without power normalization. Embedding size used in this experiment, $D_w=128$.}
\label{fig:normalization_chart}
\end{figure}

Although we apply L2 normalization and power normalization to the fisher vectors as a default choice, we study how power normalization helps in improving the results.
Fig.~\ref{fig:normalization_chart} depicts the effect of power normalization of the FVs for the document retriever step.
The chart plots top-5 accuracy for the  document retriever on the HW-SQuAD validation set, as a function of  $\lvert FV \rvert$. We use dimensionality reduced embeddings of size $D_w=128$ for different values of $K$.
It can be seen that power normalization of the FVs consistently improves the performance. The impact is more prominent as $\lvert FV \rvert$ increases. This trend is in line with the observation in previous studies~\cite{aggregate_fisher_inria,fisher_inria} that power normalization helps in improving the quality of fisher vectors for classification/retrieval problems.

Fig.~\ref{fig:propoasl_vs_accuracy} shows how the number of questions for which the target document is found within the proposals grows as a function of number of proposals. With number of proposals $>$ 2000, every question's target document is found among the proposals.

\begin{table*}[t]
\begin{center}
\begin{tabularx}{\linewidth}{X c c c c c c}
& \multicolumn{4}{c}{HW-SQuAD} &  \multicolumn{2}{c}{BenthamQA} \\
& \multicolumn{2}{c}{val} & \multicolumn{2}{c}{test} & \\

Aggregation scheme for the answer extraction step & Acc. & F1(lines)   & Acc. & F1(lines) & Acc. &F1(lines) \\ \toprule 

FV (${D_w}=128$, $K=64$) & 15.1 & 12.5  & 14.44  & 11.7  & 11.0 & 10.4  \\
FV (${D_w}=128$, $K=512$) &  10.6 & 9.9 &  & & & \\ \midrule 
SUM (${D_w}={D_o}=2048$) &  \textbf{15.9} & 12.7  & \textbf{15.9} & 12.7  & \textbf{17.5} & 15.1 \\ \bottomrule 
\end{tabularx}
\end{center}
\caption{Performance evaluation of end-to-end answer snippet extraction. In all cases, document retriever generates 5 document proposals. Aggregation scheme used for document retriever is FV (${D_w}=128$, $K=512$). Acc. stands for the snippet extraction accuracy in percentage based on DIS score proposed in Section 4.3 for a DIS threshold of 0.8. F1 (lines) is F1 score in percentage for the text lines.}
\label{table_two_stage}
\end{table*}

\subsection{Evaluating end-to-end answer snippet extraction}
\label{exp_qa}
\begin{figure*}[t]
\begin{center}
    
\end{center}
\setlength{\tabcolsep}{10pt}
\renewcommand{\arraystretch}{1}
\begin{tabularx}{\linewidth}{m{0.47\linewidth}  m{0.47\linewidth}   }

\small{\fontfamily{qhv}\selectfont When is the oldest recorded incident of civil disobedience? } &
\small{\fontfamily{qhv}\selectfont  Why was the Merit network formed in Michigan? } 
 \\

& \\

\includegraphics[height=1cm,width=8cm,cfbox=green 1pt 0pt]{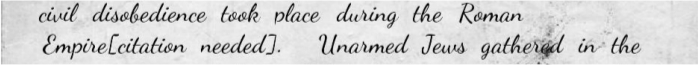} & \includegraphics[height=1cm,,width=8cm,cfbox=red 1pt 0pt]{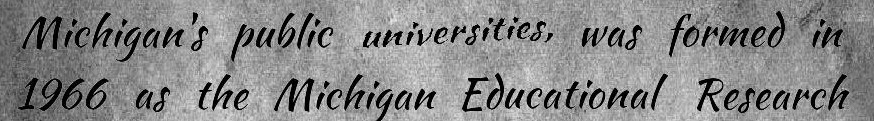}  \\

& \\
\midrule
\small{\fontfamily{qhv}\selectfont  In every quasi jury, how many classes of quasi jurors are there? } &
\small{\fontfamily{qhv}\selectfont What caused scarcity of wheat in public stores towards the end of the year 1799? } 
 \\

& \\

\includegraphics[height=1cm,width=8cm,cfbox=green 1pt 0pt]{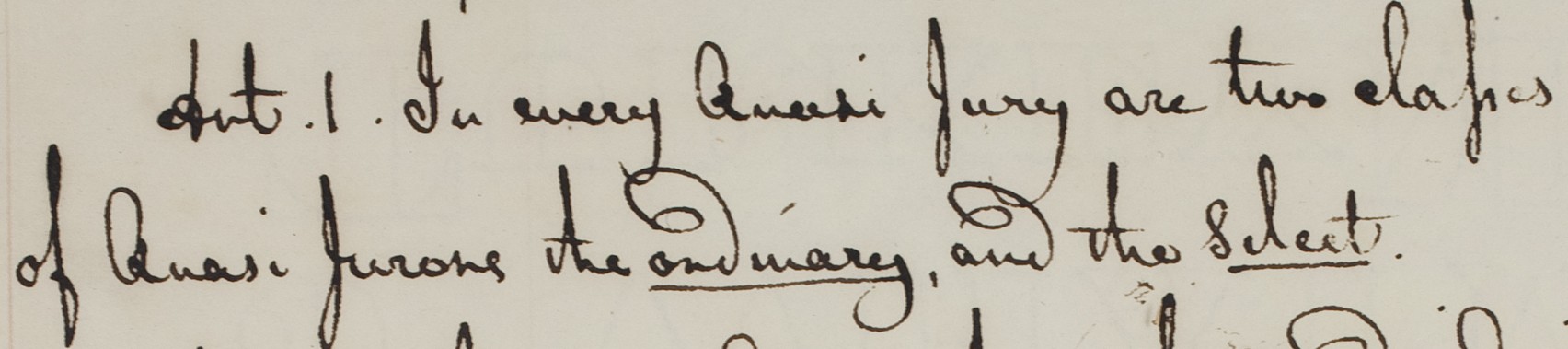} & \includegraphics[height=1cm,,width=8cm,cfbox=red 1pt 0pt]{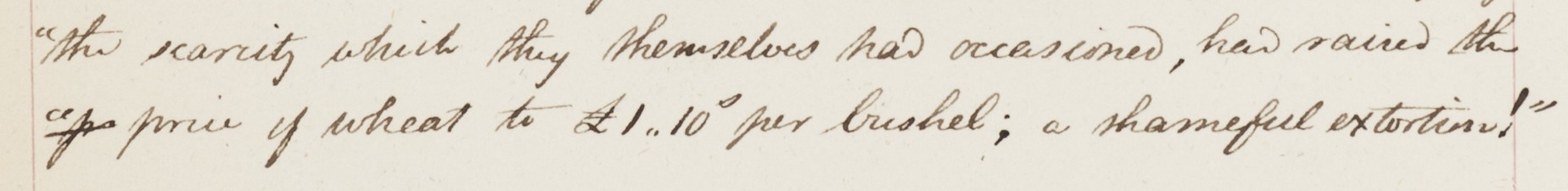}
\end{tabularx}
\caption{Qualitative Results of end-to-end answer snippet extraction. The first row shows a success and failure case from test set of HW-SQuAD. In case of the second question (the one on the right) although the extracted snippet talks about formation of Merit network it doesn't answer the question. The second row shows examples from BenthamQA dataset. The left one is a success. In case of the right one, our model returns a snippet talking about  wheat scarcity but it doesn't say anything about the reason for the scarcity.}

\label{fig:qualitative_results}
\end{figure*}

Next, we evaluate our full pipeline, which retrieves relevant documents first and then extract the answer snippet. Qualitative results of snippet extraction using the full pipeline is shown in Fig.~\ref{fig:qualitative_results}.

Table 4 shows the results for different aggregation schemes used for the snippet extraction step with our best setting for document retriever.
In addition to the accuracy measure using DIS, we use a metric that measures F1 score at the level of text lines.

Since our approach generates candidate snippets as a set of two consecutive text lines, we believe it is meaningful to define a metric at the level of text lines. We define Precision (P) as the ratio of number of text lines common to the answer snippet and the target answer to number of text lines in the answer snippet. Recall (R) is the ratio of number of lines common to the answer snippet and target answer to the number of text lines in the target answer. F1 is computed as Harmonic mean of P and R.
This metric makes sense only when the answer snippets are defined in terms of text lines. This is the reason why we do not define this metric in Section 4.3 where we define a generic evaluation scheme for the proposed QA task where answers are document snippets.

We  evaluate performance on the test set of HW-SQuAD, when the target document for each question is made available to the answer extraction step directly, bypassing the proposal generation step. This is equivalent to a MRC task since the document containing the answer is given, and the only job left is to extract the answer snippet. In this case, using the SUM aggregation, answers are found  with an accuracy of 43\%.

\begin{figure}[t]
\centering
\includegraphics[width=0.85\linewidth]{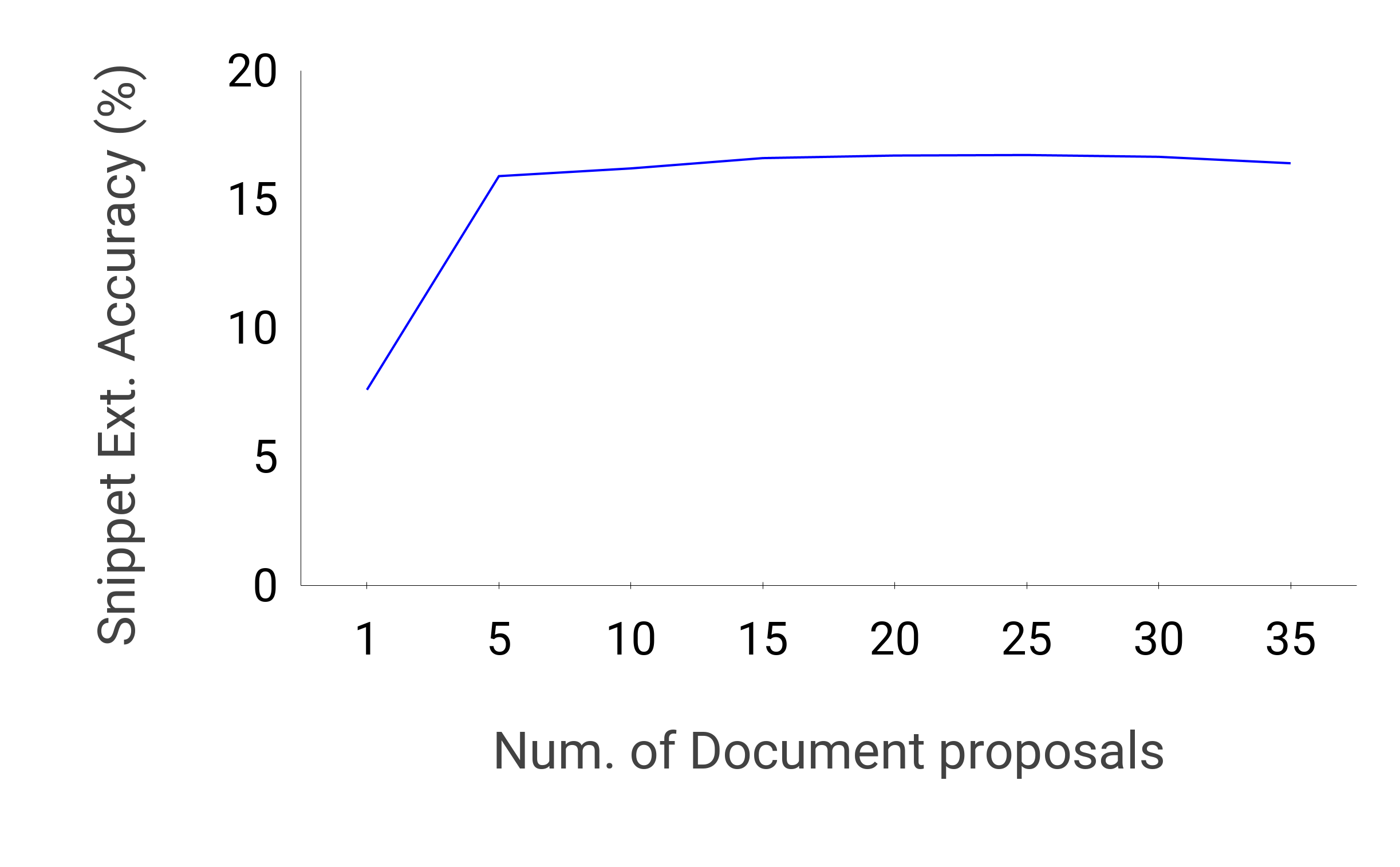}
\caption{Performance of answer snippet extraction step on the test set of HW-SQuAD as a function of number of document proposals generated. The performance flattens for number of proposals $>$ 20, suggesting that with more proposals answer extraction becomes difficult.}
\label{fig:how_many_proposals}
\end{figure}

Fig.~\ref{fig:how_many_proposals} shows how the  performance varies when the number of proposals generated varies, for the test set of HW-SQuAD.
When only one document proposal is generated, snippet extraction accuracy is 7.6\% and it goes upto 16.72\% when number of proposals is 25 and then starts dropping. When all the documents in the test set are considered as the proposals for every question in the set, the accuracy is 12.7\%.

\begin{figure}[b]
\centering
\includegraphics[width=0.85\linewidth]{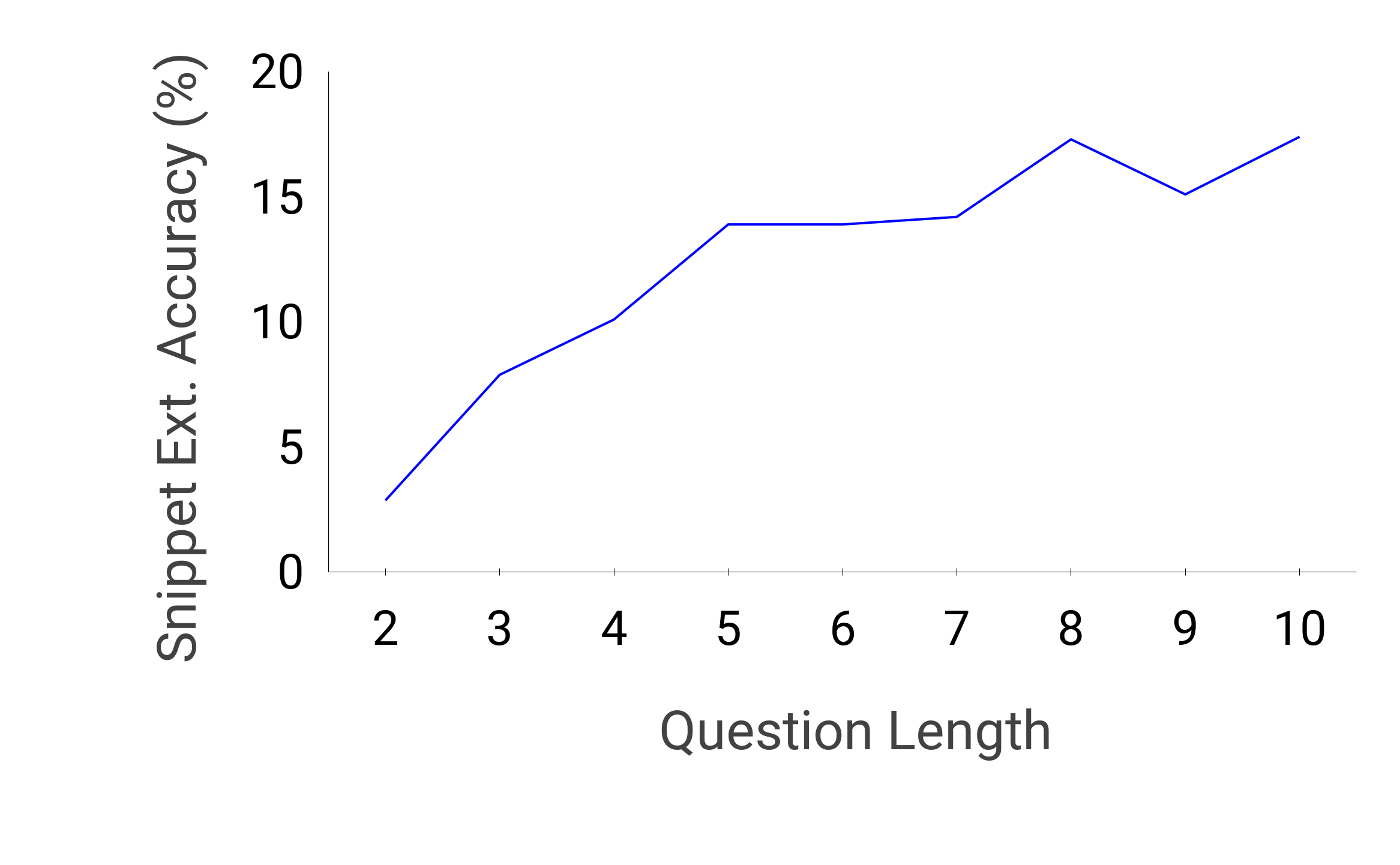}
\caption{Longer questions are likely to carry more information helpful in finding the right answer snippet when there are multiple candidates with similar content. Better performance for longer questions substantiate this.}
\label{fig:questionL}

\end{figure}

Chen et al.~\cite{reading_wiki_acl} note that open domain QA performance on SQuAD1.0 dataset is significantly affected by the nature of questions. The dataset is originally curated for MRC and the questions are asked in the context of a short paragraph. Information sought in the question is ambiguous in many cases if the context is not given. For example, questions like \textit{What day was the game played on?} or \textit{Why did he walk?} which are part of the development set of SQuAD1.0 (the test of HW-SQuAD) cannot be answered unless the associated passage or document is given. In order to study the impact of such questions on QA on HW-SQuAD, we analyze in Fig.~\ref{fig:questionL} how the performance varies when the question length varies (question length is measured in terms of the number of non stop words). The trend seen in the plot is that performance improves with more words in the question. This could mean that, with more words in the question, questions become less ambiguous and hence easier to locate the right answer snippet.

In our experiments two aspects make the experimental setting (see Section 5.1) advantageous for HW-SQuAD, compared to the BenthamQA dataset. Firstly, the off-the-shelf, end-to-end embedding model which we use for word embeddings is trained on HW-SYNTH dataset whose synthetic word images are similar to the word images in HW-SQuAD dataset.
Both HW-SYNTH and HW-SQUAD use handwritten fonts to render text and there could be many common fonts between the two datasets.
Secondly, PCA rotation matrices and GMMs used for the FV aggregation scheme are learnt on the train split of the HW-SQuAD dataset alone.
Despite the disadvantages, the document retriever and the answer extraction steps perform better on BenthamQA. We attribute this to i) the smaller size of the dataset which makes the proposal generation easier with lesser number of distractors, and ii) longer, unambiguous questions in BenthamQA as discussed in Section 4.2.

\subsection{Evaluating a recognition-based QA approach on HW-SQuAD and BenthamQA}
\label{subsec:recognition_based}
To study how recognition-based QA models work on the newly introduced Handwritten QA datasets, we evaluate a full pipeline IR/NLP QA framework on the text transcriptions of the datasets. In NLP/IR domain, the standard approach to QA over a document collection is to use a document retriever first and then use a document reader which extracts answers. The document retriever is similar in function to the one we use, i.e., to generate document proposals for the document which is likely to contain the answer. The document reader, typically an extractive QA model, extracts an answer using the document proposals as its context. In our experiments we use a TF--IDF-based document retriever, the same as the one used by Chen et al.~\cite{reading_wiki_acl} in their work. They observe that a simple TF--IDF based retriever performs better than other similar models such as ElasticSearch~\cite{elastic_search}. 
For document reader part, we use a BERT~\cite{bert} QA model. The specific model we use is a BERT\textsubscript{LARGE} model fine-tuned for QA on SQuAD1.0 dataset. The exact model name in Transformers model zoo\footnote{\url{https://huggingface.co/transformers/pretrained_models.html}} is `bert-large-uncased-whole-word-masking-finetuned-squad'. Note that HW-SQuAD which is a derivative of  SQuAD1.0 has little domain gap with HW-SQuAD data, provided the text transcriptions are good. For our experiments, we used end-to-end QA pipeline implementation available as part of the Haystack library\footnote{\url{https://github.com/deepset-ai/haystack}}.

\begin{table*}[h]
\begin{center}
\begin{tabularx}{\linewidth}{XX c c }
 && \multicolumn{2}{c}{Word acuuracy(\%)} \\
OCR & Training data& HW-SQuAD test & BenthamQA  \\ \toprule 

 SynthIam & HW-SYNTH~\cite{synth-data}+IAM~\cite{iam}   & 44.87  & 13.72 \\
SynthIamHwSq & HW-SYNTH+IAM+HW-SQuAD  & 97.85 & 23.17 \\ \bottomrule
\end{tabularx}
\end{center}
\caption{Performance of a CRNN~\cite{crnn} OCR model on the documents in the  newly introduced datasets.}
\label{table_ocr_accuracy}
\end{table*}

\subsubsection{Transcribing handwritten images using a CRNN OCR}
To recognize text in the handwritten documents in HW-SQuAD and BenthamQA, we used a CRNN~\cite{crnn} OCR model. Here the OCR model is only a word recognition model and we assume that segmented words are available. Since gold-standard word bounding boxes are available for both the the datasets, we use them directly to crop the word images. We trained two OCRs: i) SynthIam -- trained on 9 million synthetic handwritten word images in HW-SYNTH~\cite{synth-data}, train split of IAM dataset with real handwritten images, and  ii) SynthIamHwSq -- trained on HW-SYNTH, IAM train split and train split of HW-SQuAD. Except for different training data, both the OCRs are the same in terms of network architecture and training parameters.

In Table 5, we show the performance of both the OCRs on the test set of HW-SQuAD and BenthamQA.
We report word accuracy, which is the percentage of words for which the recognized text matches exactly with the ground truth. Although SynthIam is not trained on HW-SQuAD or BenthamQA, this model yields better performance on HW-SQuAD, than its performance on BenthamQA. Since the 9 million word images in HW-SYNTH are synthetically generated using handwriting fonts, it is likely that a lot of these fonts are used to render document images in HW-SQuAD. On BenthamQA although both OCRs yield low word accuracy, SynthIamHwSq is significantly better than SynthIam. We believe better performance with SynthIamHwSq is primarily due to the paper/manuscript style backgrounds used while rendering document images in HW-SQuAD. This is in contrast to IAM and HW-SYNTH where backgrounds are solid colors.
Some of the background images and textures used for HW-SQuAD resemble the background of historical manuscripts. This information must have helped the SynthIamHwSq model to better recognize the word images in BenthamQA.
\subsubsection{Evaluating TF--IDF  document retriever}
Table~\ref{table_tfidf} shows the performance of the TF-IDF document retriever for different types of transcriptions.
The results are  directly comparable with our recognition-free document retriever (Section~\ref{method_doc_retreivl}) since the task and evaluation metrics are the same. As expected, with higher OCR accuracy, the recognition-based approach is better at retrieving documents.
For BenthamQA, the proposed recognition-free approach outperforms (top-5 accuracy of 55.5\%, see Table~\ref{table_doc_fisher_full}) the recognition-based document retriever (top-5 accuracy of 32.0\%, see Table~\ref{table_tfidf}) when OCR transcriptions are used.

\begin{table}[b]
\begin{center}
\begin{tabularx}{\linewidth}{X c c }
 & \multicolumn{2}{c}{top-5 Accuracy(\%)}  \\
Transcriptions & HW-SQuAD test & BenthamQA  \\ \toprule 

 SynthIam &  47.93  & 15.76 \\
SynthIamHwSq & 86.10 & 32.00 \\ \midrule
Gold-standard  & 90.2 & 98.5 \\ \bottomrule
\end{tabularx}
\end{center}
\caption{Results of a TF--IDF document retriever on transcriptions of the documents in HW-SQuAD and BenthamQA.}
\label{table_tfidf}
\end{table}

\subsubsection{Evaluating TF--IDF + BERT\textsubscript{LARGE} full pipeline QA framework }
Here we present results of the full QA pipeline, which uses the TF--IDF document retriever to generate document proposals and a BERT\textsubscript{LARGE} based extractive QA model. The results are shown in Table~\ref{table_haystack}. In the table ``F1'' stands for the F1 score which is the most commonly used evaluation metric for QA and MRC benchmarks like SQuAD1.0. It measures the average overlap between the predicted answer and the ground truth answer~\cite{squad}. Do not confuse this F1 score with the F1 score for text lines we compute to evaluate snippet extraction in Table 4. Exact Match (EM) is the percentage of questions for which an answer matches exactly with at least one of the ground truth answers. The predicted answers are mapped to a corresponding  document snippet and evaluated for snippet extraction performance as well. Since the predicted answers are extracted as spans from the given documents, we take the text lines where the span lies as the answer snippet and evaluate by calculating the DIS score as given in Section~\ref{subsec:evaluation}.
These numbers are shown under ``Snippet Acc.'' in the table.
Comparing the snippet extraction accuracy in Table~\ref{table_two_stage} with the Snippet Acc. in Table~\ref{table_haystack}, it is evident that the proposed recognition-free approach works better when robust OCR transcriptions are not available.

\begin{table*}[tp]
\begin{center}
\begin{tabularx}{\linewidth}{X c c c c c c }
& \multicolumn{3}{c}{HW-SQuAD test} &  \multicolumn{3}{c}{BenthamQA} \\
Transcription &  F1 & EM & Snippet Acc. & F1 & EM &  Snippet Acc.  \\ \toprule 

SynthIam &  23.05 &  13.54 & 19.13  & 1.05 &  0.34 &  0.82\\
SynthIamHwSq & 65.24 & 55.31 &  59.26 & 3.00  & 1.5  & 2.46 \\ \midrule
Gold-standard  & 76.82 & 70.73 &  74.76  & 78.41 &  66.00 & 72.85 \\ \bottomrule
\end{tabularx}
\end{center}
\caption{F1 score, Exact Match (EM) percentage  and Snippet Accuracy (Snippet Acc.) for full pipeline QA using the transcribed text. F1 is the standard evaluation metric used for evaluating extractive QA. For calculating Snippet Acc., we map the textual answers to document snippets and evaluate them using DIS score proposed in Section~\ref{subsec:evaluation}. }
\label{table_haystack}
\end{table*}

Results in Table~\ref{table_haystack} suggest that highly noisy OCR transcriptions lead to sub-standard QA performance even when state-of-the-art NLP/IR QA models are used.
Although some works address recognition-based document retrieval in presence of noisy OCR outputs~\cite{venu_hw_document_retrieval_noisy_ocr,fatachia}, we do not explore this direction since our approach focuses on QA without explicit recognition of the content.

\section{Conclusion}
In this paper, we have introduced the problem of QA on handwritten document collections and presented two new datasets --- HW-SQuAD and BenthamQA.
The problem has proven to be challenging and we expect it to attract the interest of the research community.

We also have presented a new method for QA that is segmentation based, recognition-free, and lexicon free. We have presented extensive experiments with different aggregation schemes that allows us  to find answers as nearest neighbours of a given question representation in the space of possible answer snippets. The obtained results demonstrate that the proposed solution would perform better than recognition-based document retrieval and QA models, when the OCR transcription is highly noisy.
\bibliographystyle{spmpsci}
\bibliography{egbib}

\end{document}